# A short review on graphonometric evaluation tools in children.

Belen Esther Aleman, Moises Diaz, and Miguel Angel Ferrer

**Abstract**

Handwriting is a complex task that involves the coordination of motor, perceptual and cognitive skills. It is a fundamental skill for the cognitive and academic development of children. However, the technological, and educational changes in recent decades have affected both the teaching and assessment of handwriting. This paper presents a literature review of handwriting analysis in children, including a bibliometric analysis of published articles, the study participants, and the methods of evaluating the graphonometric state of children. The aim is to synthesize the state of the art and provide an overview of the main study trends over the last decade. The review concludes that handwriting remains a fundamental tool for early estimation of cognitive problems and early intervention. The article analyzes graphonometric evaluation tools. Likewise, it reflects on the importance of graphonometric evaluation as a means to detect possible difficulties or disorders in learning to write. The article concludes by highlighting the need to agree on an evaluation methodology and to combine databases.

## 1. Introduction

Handwriting is a complex skill that develops during childhood and involves coordination between sensory, motor, and cognitive systems [12]. The evaluation of handwriting is crucial in both clinical and educational fields, as it can reveal information about the neuromotor and cognitive state of the individual, detect alterations or difficulties in these systems, and evaluate the learning and teaching methods of handwriting [16]. Graphonometric analysis is a useful tool in clinical and educational contexts for diagnosing developmental or learning disorders, such as dysgraphia, which affects the process and product of handwriting [67]. Is it also useful for monitoring the evolution and recovery of patients with brain or neuromuscular injuries [25] and adapting or developing strategies to facilitate the learning of writing [7]. The objective of this paper is to present the current state of graphonometric analysis in children over a 10-year period.

This review article is divided into the following sections. The Section 2 provides a bibliometric analysis of the articles in the literature, while Section 3 examines the age of the participants and the trend in the studies. Section 4 discusses the evaluation methods of each article divided into 4.1 Objective evaluation methods, 4.2 Subjective evaluation methods and 4.3 Objective and subjective evaluation methods. Finally, Section 5 closes the manuscripts with the conclusions.

## 2. Bibliometric analysis

This section includes the bibliometric analysis of the articles published in the last 10 years on graphonometric evaluation of children handwritten. The objective of this analysis is to identify the trends, patterns and the most relevant authors included in this sample. In total, 77 articles published between 2013 and 2023 have been analyzed. It has been tried that the selected articles represent different studies within the study area. The sample was obtained from the Scopus, IEEE Xplore and Google Scholar databases using the concepts "handwriting in children", "handwriting evaluation children" and "method evaluation handwriting children". In addition, the International Graphonomic Society (IGS) proceedings of the years 2013, 2015, 2017 and 2021 have been consulted. The variables have been extracted from the select papers: number of articles published per year, authors, institutions, countries, and journals.

Fig. 1 shows the number of articles published per year on handwriting in children between 2013 and March 2023. It is observed that the average number of papers is approximately seven with peaks on the odd years corresponding to the IGS, indicating interest in this area.

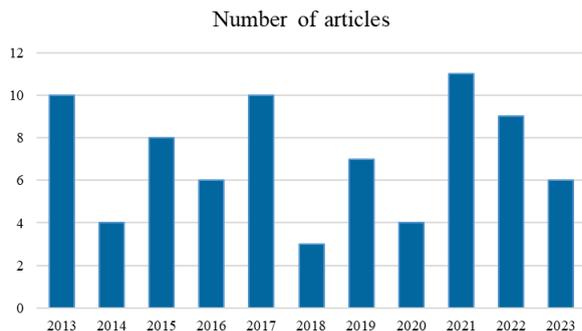

*Fig. 1.* *Number of publications per year until March 2023.*

Fig. 2 shows a network of authors and their collaboration, where the 172 authors of the different articles are gathered and interconnected. These networks are a useful tool to simplify the analysis of the degree of collaboration, the influence, productivity, internationalization, and the researchers engaged on, understanding how knowledge about handwriting develops in children.

The size of the nodes in the graph are proportional to the number of published articles in which the author has appeared. The thickness of each edge that connects the different authors is proportional to the number of collaborations in the various articles included in this work.

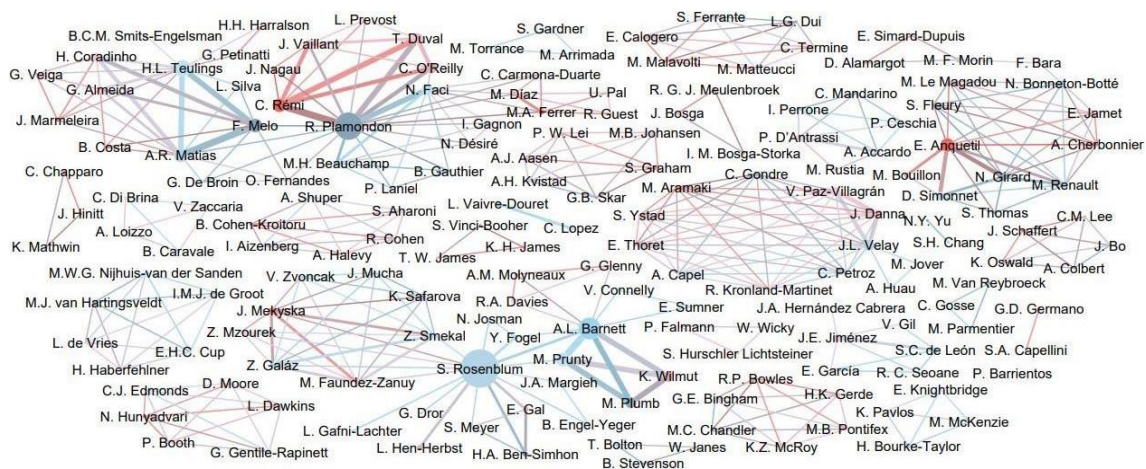

*Fig. 2.* *Author collaboration network of articles revised in this contribution.*

Upon analysis of the different papers, it was found that a total of 87 institutions have collaborated on different studies related to handwriting in children, as evidenced by the 77 articles included in the review. Figure 3 shows the 10 most repeated institutions of the 87 that have investigated handwriting in children and Figure 4 shows a map with the countries that have published more articles among those included in the sample.

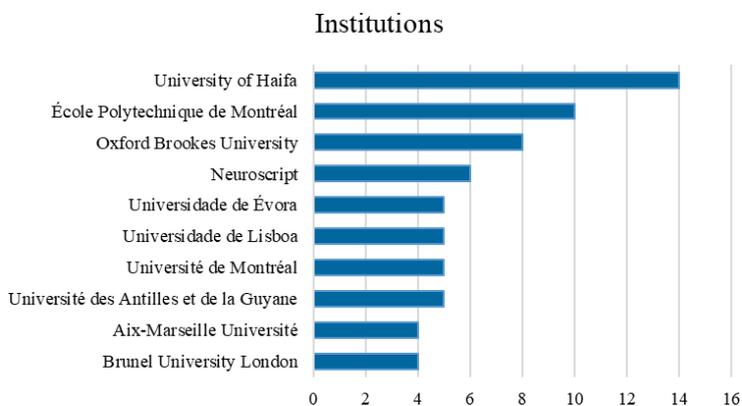

*Fig. 3.* *Bar chart of the 10 institutions with more published articles.*

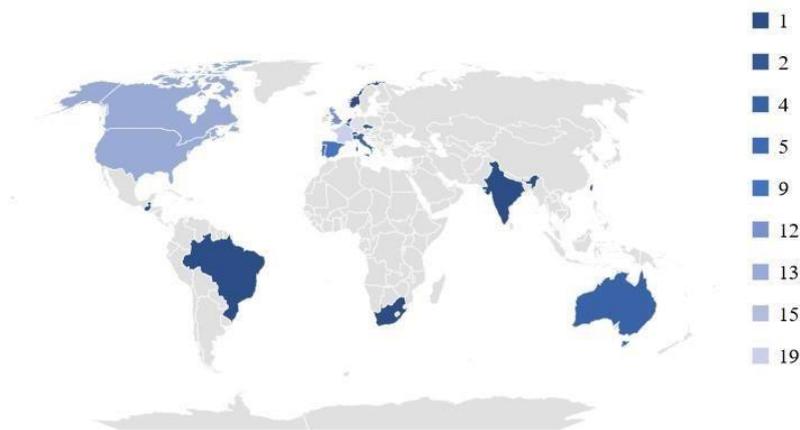

***Fig. 4.*** *Organizations and countries with the highest number of publications.*

Analyzing Fig. 3, the University of Haifa is the one that is most repeated, with 14 papers from the 87 institutions in the articles of the last 10 years included in the sample. However, Fig. 4 shows how France is the country that has led research on hand- writing in children with 19 articles, followed by Israel with 15 papers. It is also observed that Canada, the United States, the United Kingdom, and Spain are relevant in graphonometric research in children, although of these countries only two institutions from Canada, one from the United States and two from the United Kingdom appear in Figure 3.

Finally, Fig. 5 shows the 10 journals with the highest number of articles published out of the different articles analyzed. Note that the different articles analyzed have been presented in conferences, books, and journals, highlighting biennial conferences of the IGS. In this analysis, only articles that have been published in Peer review journals have been considered.

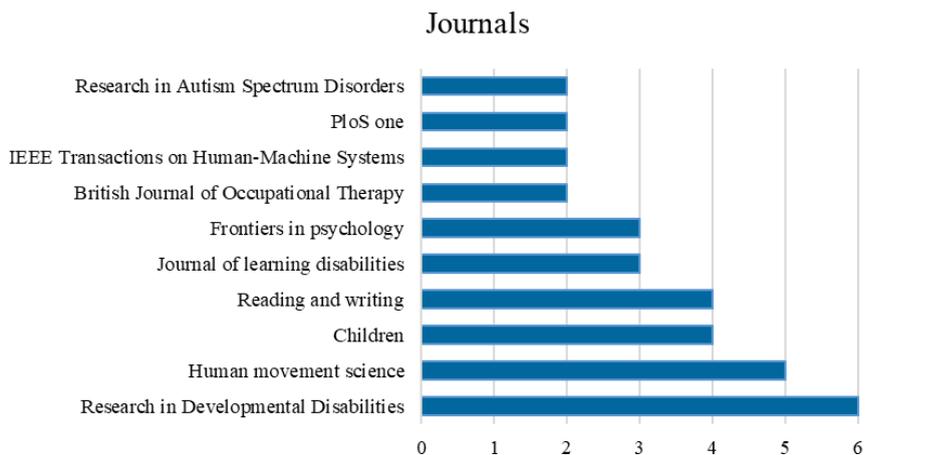

***Fig. 5.*** *Bar chart of the 10 journals with more published articles.*

### 3. Participants

The children participating in the studies of the different papers we have collected have an age range between 3 and 18 years. In Danna et al. (2013), Plamondon et al. (2013) and Paz-Villagrán et al. (2014) included children and adults in their studies, the ages of these adult participants are not included in Figure 5 as the review focuses only on graphonometry in children.

Fig. 6 shows that the studies conducted have focused on children between the ages of 6 and 12, with 9-year-olds standing out. After the age of 12, a notable decrease is observed in the studies that include participants between 13 and 18 years of age, on the other hand, as the ages of 3 to 5 years increase, the article number also rises. Therefore, there is a clear increase in the studies carried out from 3 years of age until reaching the peak at 9 years and from this age a decrease, highlighting 18 years as the age with the least studies in the articles of the sample.

Upon analysis of the data, it was observed that the studies primarily focus on evaluating handwriting in children during their primary education, with less emphasis on those who have moved beyond this stage. Additionally, it was found that children in kindergarten are predominantly the subject of evaluation in their final year before transitioning into primary education with tests according to their educational level.

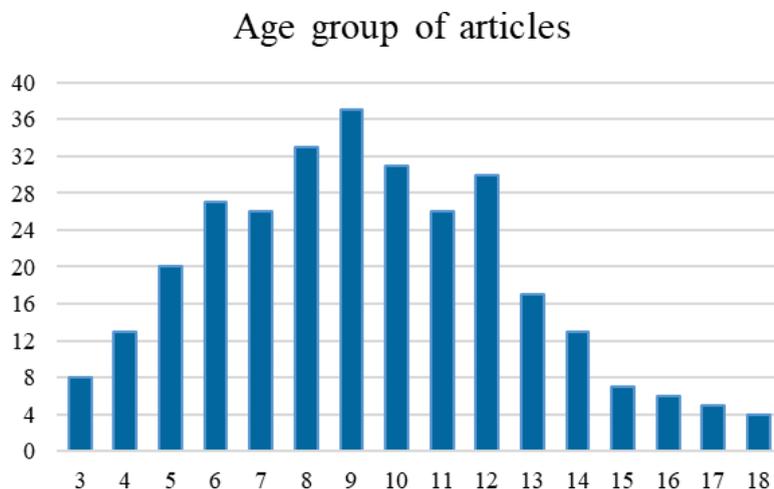

*Fig. 6.* Bar chart of age group of the study participants.

## 4. Methods of evaluation of the graphonometric state in children

A classification method based on the type of evaluation performed in each study is presented in this paper. The articles have been categorized as objective, subjective, and objective-subjective evaluation methods. The objective method pertains to evaluations where only software has been utilized for graphonometric evaluation/analysis. The subjective method refers to those articles where the evaluation has been carried out through tests in which human evaluators intervened in their evaluation. Finally, the objective and subjective method pertains to articles in which both software and standardized tests provided by evaluators were employed for evaluation.

### 4.1. Objective evaluation methods.

The objective methods present the articles that have evaluated the graphonomy of children only through software. Some studies used in their tasks the standardized tests of Concise the Assessment Scale for Children's Handwriting (BHK) [46], Detailed Assessment of Speed of Handwriting (DASH) [5], Early Grade Writing Assessment (EGWA) [4] and the figure drawing test Beery–Buktenica Developmental Test of Visual-Motor Integration (VMI) [46, 47] but these were not used to assess according to the assessment instructions. These tasks were evaluated by means of software, which collected the data and later the kinematic theory among others was evaluated. In [18] a new variable is proposed, the Signal-to-Noise velocity peaks difference (SNvpd) together with the variables number of inversion of velocity (NIV) and the averaged normalized jerk (ANJ) to calculate the fluency of handwriting in children with dysgraphia. The sigma-lognormal model ($\Sigma\Lambda$) is used to evaluate in numerous studies [20, 22, 23, 24, 25, 46, 51, 55, 60, 61] that objective measures through this model.

In the different papers revised in Table 1, the sigma-Lognormal model has been used in the most of them. This model parameterizes the movement following the kinetic theory of fast human movements. This may indicate the subject's ability to control fine motor skills approaching lognormality. In Bouillon and Anquetil (2015) pre- sent IntuiScript, a digital handwriting book project to support teaching that allows the teacher to customize the exercises according to the child's difficulties [10] and making it possible to benefit from instant feedback [70].

In most of the papers the number of participants is less than 100 participants. Among the different tasks proposed, the writing tasks designed for each study stand out. The evaluation of the different papers has evaluated quantitative measures of writing [51, 56, 57] among others, but highlights the Sigma-Lognormal parameters [20, 22, 23, 25, 55, 60, 61].

*Table 1. Manuscripts proposing objective evaluation methods.*

**Table 1.** Manuscripts proposing objective evaluation methods.

| Article | N | Tasks | Evaluation |
|---|---|---|---|
| Danna et al. (2013) [18] | 64 | Write 'lapin' | SNvpd, NIV and ANJ |
| Duval et al. (2013) [22] | 66 | Write patterns | Sigma-Lognormal |
| Molyneaux et al. (2013) [50] | 98 | Handwriting exercises | Letters, word length and frequency |
| Plamondon et al. (2013) [55] | 15 | Write patterns, drawing | Sigma-Lognormal |
| Prunty et al (2013) [56] | 56 | Five tasks from DASH | Duration, speed, execution and pause |
| Paz-Villagrán et al. (2014) [53] | 81 | Write 'lapin' | Handwriting performances |
| Prunty et al. (2014) [57] | 56 | Free writing from DASH | Handwriting pauses |
| Bouillon & Anquetil (2015) [10] | 1000 | Writing exercises | IntuiScript |
| Rémi et al. (2015) [61] | 60 | Draw scribbles | Classical dynamic and $\Sigma\Lambda$ set |
| Duval et al. (2015) [23] | 48 | Write patterns | Classical dynamic and $\Sigma\Lambda$ set |
| Vinci-Booher et al. (2016) [75] | 20 | Write letters and shapes | Functional connectivity of the brain. |
| Barrientos (2016) [4] | 120 | EGWA | Dynamics movement |
| Rosenblum & Dror (2016) [66] | 99 | Write, drawing | Dysgraphia |
| D'Antrassi et al. (2017) [17] | 257 | Draw | Kinematic parameters |
| Girard et al. (2017) [31] | 100 | Handwriting exercises | IntuiScript |
| Petinatti et al. (2017) [54] | 24 | Handwriting exercises | Dynamics movement |
| Rémi et al. (2017) [60] | | Draw doodles | Sigma-Lognormal |
| Simonnet et al. (2017) [70] | 952 | Handwriting exercises | IntuiScript |
| Teulings & Smits-Engelsman (2017) [74] | 335 | Copy | Handwriting quality and speed |
| Simonnet et al. (2019) [71] | 231 | Handwriting exercises | Handwriting quality |
| Díaz et al. (2019) [20] | 15 | Copy | Sigma-Lognormal |
| Bonneton-Botté et al. (2020) [7] | 233 | Copy | Spatiotemporal characteristics |
| Faci et al. (2021) [25] | 32 | Draw strokes | Neuromotor system integrity by $\Sigma\Lambda$. |
| Lopez & Vaivre-Douret (2021) [43] | 70 | Loops | Postural, gestural, spatial-temporal, and kinematic parameters. |
| Faci et al. (2022) [24] | 780 | Draw triangles | Sigma-Lognormal |
| Matias et al. (2022) [46] | 96 | VMI-6, BHK | Sigma-Lognormal |
| Matias et al. (2022) [47] | 110 | VMI | Process variables |
| O'Reilly et al. (2022) [51] | 780 | Draw triangles | Sigma-Lognormal |
| Germano & Capellini (2023) [29] | 95 | Write words | Latency, gaze, movement duration, fluency. |

$N^*$ denotes the number of participants in each study.

### 4.2. Subjective evaluation methods

The subjective methods present articles assessing children's graphonomy only through rater-administered assessments. The articles in the Table 2 have been evaluated on tasks with BHK [2, 42], DASH [3, 28, 58], EGWA [38], Handwriting Legibility Scale (HLS) [3], Head-Toes-Knees-Shoulders (HTKS) [11], Instructional activities for earlywriting improvement (IAEWI) [40], Indicadores de Progreso de Aprendizaje en Escritura (IPAE) [30, 40], Just Write! (JW) [6], Perceive, Recall, Plan and Perform (PRPP) [45], Standardized Test for the Evaluation of Writing withthe Key-board (TEVET) [38], VMI [6] and Wechsler Objective Language Dimensions (WOLD) [45].

The Movement Assessment Battery for Children-2 (MABC-2) [58] was included to assess the motor status of the participants. These tests incorporate some tasks and are evaluated according to the instructions of each evaluation. In addition, some articles have provided tasks that have been evaluated by several of the standardized assessments or complemented with other evaluation parameters.

*Table 2. Manuscripts proposing subjective evaluation methods.*

| Article | N | Tasks | Evaluation |
|---|---|---|---|
| Bara & Morin (2013) [2] | 332 | BHK | Handwriting style and speed, BHK |
| Prunty et al. (2016) [58] | 56 | Free writing from DASH | WOLD, DASH, MABC-2 |
| Jiménez (2017) [38] | 1653 | EGWA | EGWA comparing with TEVET |
| Barnett et al. (2018) [3] | 150 | Free writing from DASH | HLS |
| Cohen et al. (2019) [13] | 49 | Write a story. | Graphological analysis |
| Bolton et al. (2021) [6] | 37 | JW, VMI | JW comparing with VMI |
| Chandler et al. (2021) [11] | 738 | HTKS, write | Fine motor skills, HTKS, performance on writing tasks |
| Gil et al. (2021) [30] | 231 | IPAE, EGWA | IPAE, EGWA |
| Pavlos et al. (2021) [52] | 50 | HKWSA-V2, VMI | HKWSA-V2 |
| Skar et al. (2021) [72] | 4950 | Copy, write. | Writing fluency and quality |
| Fogel et al. (2022) [28] | 148 | DASH | HLS |
| Jiménez et al. (2022) [40] | 164 | IPAE, IAEWI | Teacher knowledge, intervention |
| Loizzo et al. (2023) [42] | 562 | BHK | BHK |
| Mathwin et al. (2023) [45] | 10 | Write the alphabet | PRPP |

N* denotes the number of participants in each study.

In [38] EGWA is studied, a new method of evaluation of writing in children that includes 10 copying and writing tasks. EGWA was compared with TEVET for its validation carried out by evaluators in which the results were analyzed by the theories of current writing models.

In [40] presented a level 2 intervention. The fidelity of the assessment scale (FAS) and fidelity of the intervention scale (FIS) were used. With FAS, the administration of IPAE teachers was evaluated and FIS evaluated the administration of IAEWI teachers. FAS and FIS were assessed by self-report and direct observation.

The tasks performed by the participants in most of the articles are from standardized tests, highlighting the DASH assessment, but the writing task stands out. The evaluation of the tasks does not highlight an evaluation that has been most used, in each study different aspects have been evaluated, some studies with standardized tasks were evaluated with other standardized evaluations [3, 28] and have even been evaluated by three evaluations at the same time. time [58]. The number of participants in these studies highlights one study with 4,950 [72], but most groups range from 150–738 [2,3,11,28,30,40,42].

### 4.3. Objective and subjective evaluation methods

The objective and subjective methods expose articles in which their studies were assessed both using software and peer-administered tests. As shown in Table 3, one of the software used for the evaluation was the Computerized Penmanship Evaluation Tool (ComPET) a handwriting assessment consisting of online data collection and analysis software via a pen tablet [68]. Added to previous evaluator-administered assessments, this section adds Adult Developmental Coordination Disorders/Dyspraxia (ADC) [35], Behavior Rating Inventory of Executive Function (BRIEF) [63, 64], Hebrew Handwriting Evaluation (HHE) [63, 64, 67, 68], Handwriting Proficiency Screening Questionnaire (HPSQ) [35, 49, 63, 64], (HPSQ-C) [64], Movement Assessment Battery for Children (MABC) [5, 62, 68], Minnesota Hand-writing Assessment (MHA) [5], Questionnaire for assessing students' organizational abilities-teachers (QASOA-T) [62], Lecture in a Minute (LUM) [32], Test of Visual Perceptual Skills (TVPS) [59] and World Health Organization Quality of Life Questionnaire, Brief Version (WHOQOL-BREF) [35].

In [77] children were evaluated with HPSQ-C, conventional features, and Fractional Order Derivatives (FD) based feature. FD is used as a replacement for the conventional differential derived from the extraction of the features. In this study, it was developed as a new approach for the parameterization of handwriting. With FD the basic kinematic functions (velocity, acceleration, jerk, and the horizontal and vertical variants) were extracted.

The tasks of the different studies highlight the copy tasks proposed for each study. The number of participants in most studies is in the range of 30-80 participants. Task assessment highlights the BHK [9, 19, 32, 36, 41, 44, 48] and HPSQ[35, 64, 77] assessments, as well as analysis of the handwriting process.

*Table 3. Manuscripts proposing objective and subjective evaluation methods.*

| Article | N | Tasks | Evaluation |
| --- | --- | --- | --- |
| Bosga-Storka et al. (2013) [9] | 32 | Loops, copy | BHK and Kinematic performance |
| Danna et al. (2013) [19] | 7 | Loops, copy a phrase. | Kinematic variables, BHK |
| Rosenblum et al. (2013) [68] | 58 | Copy a paragraph | Background, MABC, HPSQ, ComPET, HHE |
| Bo et al. (2014) [5] | 41 | Write letters and shapes | MABC, VMI, MHA, spatial, temporal. |
| Sumner et al. (2014) [73] | 93 | Two tasks from DASH | DASH, pause time. |
| D'Antrassi et al. (2015) [16] | 40 | Drawing, write | Qualitative and kinematic parameters |
| Huau et al. (2015) [36] | 20 | Handwriting, learning, BHK. | Spatial, spatiotemporal, dynamic variable, pen pressure, BHK |
| Rosenblum (2015) [62] | 42 | Write and copy | MABC, ComPET, QASOA-T |
| Rosenblum (2015) [63] | 64 | Copy a paragraph | HPSQ, HHE, ComPET, BRIEF |
| Rosenblum&Gafni-Lachter (2015)[67] | 230 | Copy a paragraph. | HPSQ-C, HHE, ComPET. |
| Mekyska et al. (2016) [49] | 54 | Write | HPSQ, Feature selection, intrawriter |
| Prunty et al. (2016) [59] | 56 | VMI, TVPS and DASH. | Perception and handwriting measure |
| Rosenblum et al. (2016) [69] | 60 | Write, copy. | Handwriting product and process |
| Hen-Herbst & Rosenblum (2017) [34] | 80 | Copy, write an essay. | Writing, body functions and background measures |
| Matias et al. (2017) [48] | 30 | Copy a text. | BHK and letter formation |
| Hurschler Lichtsteiner et al. (2018) [37] | 175 | Write, copy, VMI, phonological loop task | Fluency, automaticity, writing measures and intervention |
| Rosenblum (2018) [64] | 64 | Copy a paragraph. | HPSQ, HHE, ComPET and BRIEF |
| Fogel et al. (2019) [27] | 81 | Copy a paragraph. | Handwriting process, daily functions, EF |
| Jiménez & Hernández (2019)[39] | 1124 | EGWA, TEVET | EGWA, TEVET |
| Rosenblum et al. (2019) [65] | 60 | Story-writing | Production process and EF |
| Zvoncak et al. (2019) [77] | 55 | Write the Czech alphabet | HPSQ-C, conventional and FD*. |
| Alamargot et al. (2020) [1] | 45 | Write the alphabet and name. | Background measure and handwriting performance |
| Coradinho et al. (2020) [15] | 97 | VMI-6, MABC-2 | VMI-6, MABC-2, graphomotor characteristics |
| Laniel et al. (2020) [41] | 24 | Draw, BHK and Purdue Peg-board. | Intellectual functioning, graphomotor skills, BHK, neuromuscular system, behavior. |
| Bara & Bonneton-Botté(2021)[26] | 64 | Copy | Handwriting product, process, and quality |
| Dui et al. (2021) [21] | 52 | BVSCO-2 | BVSCO-2, SUS*, satisfaction, tilt, in-air time |
| Gosse et al. (2021) [32] | 117 | Chronosdictées, BHK | Chronosdictées, BHK, LUM |
| Torrance et al. (2021) [76] | 179 | Copy, write | Spelling, fluency, letters, phonetic, accuracy, reading, reasoning. |
| Booth et al. (2022) [8] | 85 | Hand tasks, write | Kinematics and handwriting quality |
| Chang & Yu (2022) [12] | 641 | Copy | Geometric, spatiotemporal measures |
| Hen-Herbst&Rosenblum(2022)[35] | 80 | Copy, WHOQOL-BREF, ADC | HPSQ, HLS, ComPET, ADC, WHOQOL-BREF |
| Coradinho et al. (2023) [14] | 57 | BHK | Handwriting product and process |
| Haberfehlner et al. (2023) [33] | 374 | Drawing | Handwriting readiness |
| Lopez & Vaivre-Douret(2023)[44] | 35 | BHK, loops. | BHK, spatial temporal and kinematic |

*N\* denotes the number of participants in each study. FD\*: Fractional Order Derivatives. SUS\* System Usability Scale*

## 5. Discussion and Conclusion

In conclusion, the evaluation of handwriting in children is a complex process that re-quires appropriate methods and instruments that must be systematic, objective, and sensitive to the different factors involved. There are various waysto evaluate hand- writing, including software and expert evaluation, each with its advantages and disadvantages. While software evaluation is fast, accurate, and objective based on pre- defined parameters, it may not capture some qualitative or contextual aspects of the written product. Conversely, expert evaluation may be more flexible and responsive to the characteristics of the written specimen but may introduce a subjective bias or evaluator fatigue, whichcould affect the reliability and validity of the results.

Furthermore, it is essential to recognize that the emotional factor can influence the written process and product, as handwriting is not only a means of communication and learning but also an expression of personality, emotions, and feelings. Thus, fac- tors such as children's self-esteem, motivation, and academic performance should also be considered in the evaluation of handwriting [12].

In [9, 25, 30, 32] longitudinal studies are carried out, these studies allow to observe the evolution in time of the handwriting of the participants in tasks. Increasing these studies with longitudinal databases would allow a better understanding of the evolution of handwriting in children and be able to apply tools for learning this skill or new methods for diagnosing different learning problems.

These tools will enable accurate and reliable evaluations, which will ultimately lead to improved interventions and outcomes for children's cognitive and academic development.

On the other hand, the different studies have seen that of graphonomic evaluation under the kinetic theory using the Sigma-Lognormal parameters in different writing and drawing tasks, evaluating the dynamic movements that these tasks imply. Different standardized evaluations have been used, but the use of some more than others stands out, such as the case of BHK and DASH. The BHK and DASH evaluations have been used in several articles, in some only their tasks were applied, and they were evaluated by other criteria. It should be noted that although these evaluations used their tasks in the subjective and objective-subjective methods, there is a lack of consensus between authors for a common task to evaluate the same aspects, especially in the papers included in objective methods.

Evaluations with human involvement and software provide different measures depending on the evaluation to be carried out and the proposed tasks. To address these challenges, it is necessary to agree on the development of an evaluation methodology that is partly common in the recording protocols, allowing faster progress by being able to combine the databases, increasing their size and making it possible to compare the different algorithms on the databases. Considering the different assessments and alphabets used in each task to better understand the way in which handwriting is taught and acquired depending on the type of alphabet and the cultural context. This could lead to a better analysis of the advantages and challenges of the systems, as well as intervention strategies to improve the learning of handwriting in different contexts.

**References**

1. Alamargot, D., Morin, M. F., & Simard-Dupuis, E. (2020). Handwriting delay in dyslexia: Children at the end of primary school still make numerous short pauses when producing letters. Journal of learning disabilities, 53(3), 163-175.
2. Bara, F., & Morin, M. F. (2013). Does the handwriting style learned in first grade deter mine the style used in the fourth and fifth grades and influence handwriting speed and quality? A comparison between French and Quebec children. Psychology in the Schools, 50(6), 601-617.
3. Barnett, A. L., Prunty, M., & Rosenblum, S. (2018). Development of the Handwriting Leg ibility Scale (HLS): A preliminary examination of Reliability and Validity. Research in developmental disabilities, 72, 240-247.
4. Barrientos, P. (2017). Handwriting development in Spanish children with and without learning disabilities: A graphonomic approach. Journal of learning disabilities, 50(5), 552 563.
5. Bo, J., Colbert, A., Lee, C. M., Schaffert, J., Oswald, K., & Neill, R. (2014). Examining the relationship between motor assessments and handwriting consistency in children with and without probable developmental coordination disorder. Research in developmental disabilities, 35(9), 2035-2043.
6. Bolton, T., Stevenson, B., & Janes, W. (2021). Assessing handwriting in preschool-aged children: Feasibility and construct validity of the "Just Write!" tool. Journal of Occupa tional Therapy, Schools, & Early Intervention, 14(2), 153-161.
7. Bonneton-Botté, N., Fleury, S., Girard, N., Le Magadou, M., Cherbonnier, A., Renault, M., ... & Jamet, E. (2020). Can tablet apps support the learning of handwriting? An inves tigation of learning outcomes in kindergarten classroom. Computers & Education, 151, 103831.
8. Booth, P., Hunyadvari, N., Dawkins, L., Moore, D., Gentile-Rapinett, G., & Edmonds, C. J. (2022). Water consumption increases handwriting speed and volume consumed relates to increased finger-tapping speed in schoolchildren. Journal of Cognitive Enhancement, 6(2), 183-191.
9. Bosga-Storka, I. M., Bosgaa, J., Meulenbroekb, R. G., Bosga-Stork, P., & Beaufortweg, D. Age-related Changes in Regularity Statistics of Loop-writing Kinematics. In Recent Progress in Graphonomics: Learn from the Past. IGS 2013 (pp. 109-114).
10. Bouillon, M., & Anquetil, E. (2015, June). Handwriting analysis with online fuzzy models. In 17th Biennial Conference of the International Graphonomics Society.
11. Chandler, M. C., Gerde, H. K., Bowles, R. P., McRoy, K. Z., Pontifex, M. B., & Bingham, G. E. (2021). Self-regulation moderates the relationship between fine motor skills and writing in early childhood. Early Childhood Research Quarterly, 57, 239-250.
12. Chang, S. H., & Yu, N. Y. (2022). Computerized handwriting evaluation and statistical reports for children in the age of primary school. Scientific Reports, 12(1), 15675.

13. Cohen, R., Cohen-Kroitoru, B., Halevy, A., Aharoni, S., Aizenberg, I., & Shuper, A. (2019). Handwriting in children with Attention Deficit Hyperactive Disorder: role of graphology. BMC pediatrics, 19(1), 1-6.
14. Coradinho, H., Melo, F., Almeida, G., Veiga, G., Marmeleira, J., Teulings, H. L., & Matias, A. R. (2023). Relationship between Product and Process Characteristics of Handwriting Skills of Children in the Second Grade of Elementary School. Children, 10(3), 445.
15. Coradinho, H., Melo, F., Teulings, H. L., Matias, A. R., & Matias, A. (2020). COMPETÊNCIA MOTORA E COMPETÊNCIAS GRAFOMOTORAS EM CRIANÇAS NO ÚLTIMO ANO DO PRÉ-ESCOLAR.
16. D'Antrassi, P., Ceschia, P., Mandarino, C., Perrone, I., & Accardo, A. (2015, June). Evaluation of Different Handwriting Teaching Methods by Kinematic and Quality Analyses. In 17th Biennial Conference of the International Graphonomics Society.
17. D'Antrassi, P., Rustia, M., & Accardo, A. (2017). Development of graphomotor skills in school-age children. In Graphonomics for e-Citizens: e-Health, e-Society, e-Education (pp. 205-208). Claudio De Stefano and Angelo Marcelli.
18. Danna, J., Paz-Villagrán, V., & Velay, J. L. (2013). Signal-to-Noise velocity peaks difference: A new method for evaluating the handwriting movement fluency in children with dysgraphia. Research in developmental disabilities, 34(12), 4375-4384.
19. Danna, J., Velay, J. L., Vietminh, V., Capel, A., Petroz, C., Gondre, C., ... & Kronland Martinet, R. (2013, October). Handwriting movement sonification for the rehabilitation of dysgraphia. In 10th International Symposium on Computer Music Multidisciplinary Research (CMMR)-Sound, Music & Motion-15-18 oct. 2013-Marseille, France (pp. 200 208).
20. Diaz, M., Ferrer, M., Guest, R., & Pal, U. (2019). Graphomotor Evolution in the Handwriting of Bengali Children Through Sigma-Lognormal Based-Parameters: A Preliminary Study.
21. Dui, L. G., Calogero, E., Malavolti, M., Termine, C., Matteucci, M., & Ferrante, S. (2021, July). Digital tools for handwriting proficiency evaluation in children. In 2021 IEEE EMBS International Conference on Biomedical and Health Informatics (BHI) (pp. 1-4). IEEE.
22. Duval, T., Plamondon, R., O'Reilly, C., Remi, C., & Vaillant, J. (2013, June). On the use of the sigma-lognormal model to study children handwriting. In Recent Progress in Graphonomics: Learn from the Past. IGS 2013 (pp. 26-30).
23. Duval, T., Rémi, C., Plamondon, R., Vaillant, J., & O'Reilly, C. (2015). Combining sigma-lognormal modeling and classical features for analyzing graphomotor performances in kindergarten children. Human movement science, 43, 183-200.
24. Faci, N., Carmona-Duarte, C., Diaz, M., Ferrer, M. A., & Plamondon, R. (2022, December). Comparison Between Two Sigma-Lognormal Extractors with Primary Schools Students Handwriting. In Intertwining Graphonomics with Human Movements: 20th International Conference of the International Graphonomics Society, IGS 2021, Las Palmas de Gran Canaria, Spain, June 7-9, 2022, Proceedings (pp. 105-113). Cham: Springer International Publishing.
25. Faci, N., Désiré, N., Beauchamp, M. H., Gagnon, I., & Plamondon, R. (2021). Analysing the Evolution of Children's Neuromotor System Lognormality after Mild Traumatic Bain Injury. In THE LOGNORMALITY PRINCIPLE AND ITS APPLICATIONS IN E SECURITY, E-LEARNING AND E-HEALTH (pp. 143-160).
26. Florence, B., & Nathalie, B. B. (2021). Handwriting isolated cursive letters in young children: Effect of the visual trace deletion. Learning and Instruction, 74, 101439.
27. Fogel, Y., Josman, N., & Rosenblum, S. (2019). Functional abilities as reflected through temporal handwriting measures among adolescents with neuro-developmental disabilities. Pattern Recognition Letters, 121, 13-18.
28. Fogel, Y., Rosenblum, S., & Barnett, A. L. (2022). Handwriting legibility across different writing tasks in school-aged children. Hong Kong Journal of Occupational Therapy, 15691861221075709. 29. Germano, G. D., & Capellini, S. A. (2023). Handwriting fluency, latency, and kinematic in Portuguese writing system: Pilot study with school children from 3rd to 5th grade. Frontiers in Psychology, 13, 1063021.
30. Gil, V., de León, S. C., & Jiménez, J. E. (2021). Universal screening for writing risk in Spanish-speaking first graders. Reading & Writing Quarterly, 37(2), 117-135.
31. Girard, N., Simonnet, D., & Anquetil, E. (2017, June). IntuiScript a new digital notebook for learning writing in elementary schools: 1st observations. In 18th International Graphonomics Society Conference (IGS2017) (pp. 201-204).
32. Gosse, C., Parmentier, M., & Van Reybroeck, M. (2021). How do spelling, handwriting speed, and handwriting quality develop during primary school? Cross-classified growth curve analysis of children's writing development. Frontiers in Psychology, 12, 685681.
33. Haberfehlner, H., de Vries, L., Cup, E. H., de Groot, I. J., Nijhuis-van der Sanden, M. W., & van Hartingsveldt, M. J. (2023). Ready for handwriting? A reference data study on handwriting readiness assessments. Plos one, 18(3), e0282497.
34. Hen-Herbst, L. & Rosenblum, S. (2017). Executive Functions, Coordination and Developmental Functional Abilities as Predictors of Writing Capabilities among Adolescents. In 18th International Graphonomics Society Conference (IGS2017) (pp. 6-10)

35. Hen-Herbst, L., & Rosenblum, S. (2022). Handwriting and Motor-Related Daily Performance among Adolescents with Dysgraphia and Their Impact on Physical Health-Related Quality of Life. Children, 9(10), 1437.
36. Huau, A., Velay, J. L., & Jover, M. (2015). Graphomotor skills in children with developmental coordination disorder (DCD): Handwriting and learning a new letter. Human movement science, 42, 318-332.
37. Hurschler Lichtsteiner, S., Wicki, W., & Falmann, P. (2018). Impact of handwriting train ing on fluency, spelling and text quality among third graders. Reading and writing, 31, 1295-1318.
38. Jiménez, J. E. (2017). Early grade writing assessment: An instrument model. Journal of Learning Disabilities, 50(5), 491-503.
39. Jiménez, J. E., & Hernández-Cabrera, J. A. (2019). Transcription skills and written composition in Spanish beginning writers: Pen and keyboard modes. Reading and Writing, 32(7), 1847-1879.
40. Jiménez, J. E., de León, S. C., García, E., & Seoane, R. C. (2022). Assessing the efficacy of a Tier 2 early intervention for transcription skills in Spanish elementary school students. Reading and Writing, 1-33.
41. Laniel, P., Faci, N., Plamondon, R., Beauchamp, M. H., & Gauthier, B. (2020). Kinematic analysis of fast pen strokes in children with ADHD. Applied Neuropsychology: Child, 9(2), 125-140.
42. Loizzo, A., Zaccaria, V., Caravale, B., & Di Brina, C. (2023). Validation of the Concise Assessment Scale for Children's Handwriting (BHK) in an Italian Population. Children, 10(2), 223.
43. Lopez, C., & Vaivre-Douret, L. (2021). Influence of visual control on the quality of graph ic gesture in children with handwriting disorders. Scientific Reports, 11(1), 23537.
44. Lopez, C., & Vaivre-Douret, L. (2023). Concurrent and Predictive Validity of a Cycloid Loops Copy Task to Assess Handwriting Disorders in Children. Children, 10(2), 305.
45. Mathwin, K., Chapparo, C., & Hinitt, J. (2023). Children with handwriting difficulties: Impact of cognitive strategy training for acquisition of accurate alphabet-letter-writing. British Journal of Occupational Therapy, 03080226221148413.
46. Matias, A. R., Melo, F., Coradinho, H., Fernandes, O., de Broin, G., & Plamondon, R. (2022, December). Effects of a Graphomotor Intervention on the Graphic Skills of Children: An Analysis with the Sigma-Lognormal Model. In Intertwining Graphonomics with Human Movements: 20th International Conference of the International Graphonomics Society, IGS 2021, Las Palmas de Gran Canaria, Spain, June 7-9, 2022, Proceedings (pp. 114-128). Cham: Springer International Publishing.
47. Matias, A. R., Melo, F., Costa, B., Teulings, H. L., Almeida, G. (2022). Copy of geometric drawings in children from 4 to 6 years old: a kinematic analysis. In 20th Conference of the International Graphonomics Society (IGS2021)
48. Matias, A. R., Teulings, H. L., Silva, L., & Melo, F. (2017). Measuring handwriting stability versus context variations. Presentation at IGS2017.
49. Mekyska, J., Faundez-Zanuy, M., Mzourek, Z., Galaz, Z., Smekal, Z., & Rosenblum, S. (2016). Identification and rating of developmental dysgraphia by handwriting analysis. IEEE Transactions on Human-Machine Systems, 47(2), 235-248.
50. Molyneaux, A. M., Barnett, A. L., Glenny, G., Davies, R. A. (2013). The Association between Handwriting Practice and Lexical Richness: An Analysis of the Handwritten Output of Children aged 9 - 10 years. In Recent Progress in Graphonomics: Learn from the Past. IGS 2013 (pp. 38-41).
51. O'Reilly, C., Plamondon, R., & Faci, N. (2022, August). The Lognometer: A New Nor malized and Computerized Device for Assessing the Neurodevelopment of Fine Motor Control in Children. In 2022 26th International Conference on Pattern Recognition (ICPR) (pp. 952-958). IEEE.
52. Pavlos, K., McKenzie, M., Knightbridge, E., & Bourke-Taylor, H. (2021). Initial psycho metric evaluation of the Hartley Knows Writing Shapes Assessment Version 2 with typically developing children between the ages of 4 and 8. Australian Occupational Therapy Journal, 68(1), 32-42.
53. Paz-Villagrán, V., Danna, J., & Velay, J. L. (2014). Lifts and stops in proficient and dys graphic handwriting. Human movement science, 33, 381-394.
54. Petinatti, G., Harralson, H. H., Teulings, H. L. (2017). Does Exercise in Children with Learning Disabilities Improve Cursive Handwriting?
55. Plamondon, R., O'Reilly, C., Rémi, C., & Duval, T. (2013). The lognormal handwriter: learning, performing, and declining. Frontiers in psychology, 4, 945.
56. Prunty, M. M., Barnett, A. L., Wilmut, K., & Plumb, M. S. (2013). Handwriting speed in children with Developmental Coordination Disorder: Are they really slower? Research in developmental disabilities, 34(9), 2927-2936.
57. Prunty, M. M., Barnett, A. L., Wilmut, K., & Plumb, M. S. (2014). An examination of writing pauses in the handwriting of children with Developmental Coordination Disorder. Research in developmental disabilities, 35(11), 2894-2905.

58. Prunty, M. M., Barnett, A. L., Wilmut, K., & Plumb, M. S. (2016). The impact of handwriting difficulties on compositional quality in children with developmental coordination disorder. British Journal of Occupational Therapy, 79(10), 591-597.
59. Prunty, M., Barnett, A. L., Wilmut, K., & Plumb, M. (2016). Visual perceptual and handwriting skills in children with Developmental Coordination Disorder. Human movement science, 49, 54-65.
60. Rémi, C., Nagau, J., Vaillant, J., & Plamondon, R. (2017, June). Preliminary study of t0, a sigma-lognormal parameter extracted from young children's-controlled scribbles. In 18th Conference of the International Graphonomics Society, IGS2017 (pp. 109-113).
61. Rémi, C., Vaillant, J., Plamondon, R., Prevost, L., & Duval, T. (2015, June). Exploring the kinematic dimensions of kindergarten children's scribbles. In 17th Biennial Conference of the International Graphonomics Society (pp. 79-82). 62. Rosenblum, S. (2015). Do motor ability and handwriting kinematic measures predict organizational ability among children with developmental coordination disorders?. Human movement science, 43, 201-215.
63. Rosenblum, S. (2015). Relationships between handwriting features and executive control among children with developmental dysgraphia. Drawing, Handwriting Processing Analy sis: New Advances and Challenges, 111.
64. Rosenblum, S. (2018). Inter-relationships between objective handwriting features and executive control among children with developmental dysgraphia. PloS one, 13(4), e0196098.
65. Rosenblum, S., Ben-Simhon, H. A., Meyer, S., & Gal, E. (2019). Predictors of handwriting performance among children with autism spectrum disorder. Research in Autism Spectrum Disorders, 60, 16-24.
66. Rosenblum, S., & Dror, G. (2016). Identifying developmental dysgraphia characteristics utilizing handwriting classification methods. IEEE Transactions on Human-Machine Systems, 47(2), 293-298.
67. Rosenblum, S., & Gafni-Lachter, L. (2015). Handwriting proficiency screening questionnaire for children (HPSQ–C): development, reliability, and validity. The American Journal of Occupational Therapy, 69(3), 6903220030p1-6903220030p9.
68. Rosenblum, S., Margieh, J. A., & Engel-Yeger, B. (2013). Handwriting features of children with developmental coordination disorder–results of triangular evaluation. Research in developmental disabilities, 34(11), 4134-4141.
69. Rosenblum, S., Simhon, H. A. B., & Gal, E. (2016). Unique handwriting performance characteristics of children with high-functioning autism spectrum disorder. Research in Autism Spectrum Disorders, 23, 235-244.
70. Simonnet, D., Anquetil, E., & Bouillon, M. (2017). Multi-criteria handwriting quality analysis with online fuzzy models. Pattern Recognition, 69, 310-324. 71. Simonnet, D., Girard, N., Anquetil, E., Renault, M., & Thomas, S. (2019). Evaluation of children cursive handwritten words for e-education. Pattern Recognition Letters, 121, 133 139.
72. Skar, G. B., Lei, P. W., Graham, S., Aasen, A. J., Johansen, M. B., & Kvistad, A. H. (2021). Handwriting fluency and the quality of primary grade students' writing. Reading and Writing, 1-30.
73. Sumner, E., Connelly, V., & Barnett, A. L. (2014). The influence of spelling ability on handwriting production: children with and without dyslexia. Journal of Experimental Psychology: Learning, Memory, and Cognition, 40(5), 1441.
74. Teulings, H. L. & Smits-Engelsman, B. C. (2017). Objective Measurement of Handwriting Learning Outcomes at Elementary schools Suggest Instruction Improvements. In 18th International Graphonomics Society Conference (IGS2017) (pp. 1-5)
75. Vinci-Booher, S., James, T. W., & James, K. H. (2016). Visual-motor functional connectivity in preschool children emerges after handwriting experience. Trends in Neuroscience and Education, 5(3), 107-120.
76. Torrance, M., Arrimada, M., & Gardner, S. (2021). Child-level factors affecting rate of learning to write in first grade. British Journal of Educational Psychology, 91(2), 714-734.
77. Zvoncak, V., Mucha, J., Galaz, Z., Mekyska, J., Safarova, K., Faundez-Zanuy, M., & Smekal, Z. (2019, October). Fractional order derivatives evaluation in computerized assessment of handwriting difficulties in school-aged children. In 2019 11th International Congress on Ultra-Modern Telecommunications and Control Systems and Workshops (ICUMT) (pp. 1-6). IEEE.